\pdfoutput=1
\documentclass[11pt]{article}

\usepackage[]{EMNLP2023}

\usepackage{times}
\usepackage{latexsym}

\usepackage{color, colortbl}
\usepackage{booktabs}
\usepackage{pifont}
\usepackage{array}
\usepackage{multirow}
\usepackage{enumitem}
\usepackage{graphicx}
\usepackage{hyperref}
\usepackage{amsmath}
\usepackage{subcaption}
\usepackage{float}
\usepackage{amsmath}

\definecolor{LightGray}{gray}{0.85}

\usepackage[T1]{fontenc}
\usepackage[utf8]{inputenc}

\usepackage{microtype}
\usepackage{upgreek}

\usepackage{inconsolata}
\usepackage{booktabs} % For formal tables

\title{ProtoMed-LLM: An Automatic Evaluation Framework for \\ Large Language Models in Medical Protocol Formulation }
\author{
Seungjun Yi\textsuperscript{1,3*}, Jaeyoung Lim\textsuperscript{2,3}, Juyong Yoon\textsuperscript{3*}\\
\textsuperscript{1} Department of Biomedical Engineering, University of Texas at Austin\\
\textsuperscript{2} Department of Computer Science and Engineering, Ulsan National Institute of Science and Technology \\
\textsuperscript{3} Korea Institute of Science and Technology (KIST) Europe \\
\{charlie.yi@utexas.edu, ljy7223@khu.ac.kr, juyong.yoon@kist-europe.de\}
}

\begin{document}
\maketitle
\begin{abstract}
Automated generation of scientific protocols executable by robots can significantly accelerate scientific research processes. Large Language Models (LLMs) excel at Scientific Protocol Formulation Tasks (SPFT), but the evaluation of their capabilities rely on human evaluation. Here, we propose a flexible, automatic framework to evaluate LLMs' capability on SPFT: \textit{ProtoMed-LLM}\footnote{The dataset and code are available \href{https://github.com/ProtocoL-LLM/ProtocoLLM.git}{here}.\\ \indent\indent*Corresponding Authors}. 
%defines domain-specific experimental actions, 
This framework prompts the target model and GPT-4 to extract pseudocode from biology protocols using only predefined lab actions and evaluates the output of target model using \textsc{Llam-Eval}, the pseudocode generated by GPT-4 serving as a baseline and Llama-3 acting as the evaluator. Our adaptable prompt-based evaluation method, \textsc{Llam-Eval}, offers significant flexibility in terms of evaluation model, material, criteria, and is free of cost.
% G-Eval 용어는 일부러 말하지 않음
% independently
We evaluate GPT variations, Llama, Mixtral, Gemma, Cohere, and Gemini. Overall, we find that GPT and Cohere is a powerful scientific protocol formulators.
We also introduce \textsc{Bioprot 2.0}, a dataset with biology protocols and corresponding pseudocodes, which can aid LLMs in formulation and evaluation of SPFT.
Our work is extensible to assess LLMs on SPFT across various domains and other fields that require protocol generation for specific goals.

\end{abstract}

% 추가할 점
% G-Eval 필요성? 간단설명 
% traditional method나 next step prediction 같은 task 만으로 protocol quality를 평가하는 것은 쉽지 않다 (bioplanner에서 보면 accuracy가 높아봤자 60프로면 완전 꽝인듯) . 

\section{Introduction}
\label{sec:introduction}

Laboratory automation is essential for accelerating scientific research processes. However, most contemporary laboratories use manual labor, especially in the field of biology. This not only constrains the scope for scalability, but also introduces potential vulnerabilities in reproducibility \citep{Kwok2010}. 

One of the barriers for automation in biology is the reliance on manual experiments when validating scientific protocols. Traditionally, trial-and-error approach has been employed to formulate a protocol to achieve a certain goal. As a breakthrough, LLMs have demonstrated remarkable capabilities in formulating precise experimental protocols across diverse fields \citep{chem_01, chem_mat}. These protocols comprise pseudocodes with actionable sequences that can be executed by machines which can be automated. Yet, efforts in biology to utilize LLMs for pseudocode formulation are yet to achieve desired outcomes \citep{Inagaki2023LLMsCG}. These works rely on human evaluations, and objective evaluation methods for protocol formulation are nonexistent. Therefore, it is necessary to establish an automated evaluation framework on formulating protocols to move beyond manual labor.

% Wording을 prediction이라고 써서 헷갈릴수도? -> 다른 wording 생각해보기
\begin{figure}[!t] % Place figure at the top of the page
  \centering % Centers the figure
  \includegraphics[width=0.4\textwidth]{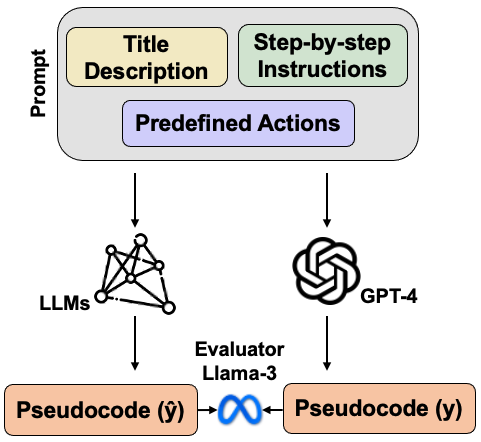} %
  \caption{\textbf{Overview of the \textit{ProtoMed-LLM} Framework.} A protocol containing a title, descriptions, step-by-step instructions, and predefined biology lab actions is given to both a target model and GPT-4 for pseudocode generation. Then, Llama-3 evaluates these outputs considering the target model's pseudocode as the prediction ($\hat{y}$) and GPT-4's as a  baseline ($y$).}
  
  %Then GPT as an evaluator is used to compare the outcomes, considering the target LLM-generated pseudocode as a prediction ($\hat{y}$) and GPT-generated pseudocode the ground truth ($y$).} % Adds a caption to the figure

  %The ProtocoLLM Framework. This framework involves providing titles, descriptions, step-by-step instructions, and a set of predefined biology lab actions to both a target model and GPT-4 for pseudocode generation. GPT then evaluates these outputs, treating the target model's pseudocode as a prediction () and its own as the ground truth (y).
  \label{fig:diagram} % Allows for referencing the figure in the text
\end{figure}
Previous work suggests a framework to assess the capabilities of LLMs on SPFT: \textit{BioPlanner} \citep{odonoghue-etal-2023-bioplanner}. This method outlines three primary steps: (i) extracting pseudofunctions and pseudocode\footnote{Pseudofunctions represent laboratory actions, while pseudocode embodies protocols composed of these pseudofunctions.} 
% Examples can be found at Appendix~\ref{sec:appendix-exampledata}.
from a protocol using an evaluator, (ii) using the target model to produce pseudocode given the pseudofunctions, and (iii) evaluating the pseudocode generated in step (ii) against the original pseudocode in (i). Using this framework, they performed evaluation exclusively on GPTs~\citep{brown2020language, openai2023gpt4}. 

We highlight the following key observations:
(1) Various representations of pseudofunctions corresponding to identical experimental actions, causes performance degradation and inconsistency of the evaluation framework.
% performance degradation 참고문헌 추가 또는 appendix
(2) The repertoire of actions executed in biology labs is confined to a finite set of actions. 
% 3번 수정 필요
(3) High values in traditional automatic metrics~\hyperref[LLMeval:1]{(i)} does not necessarily imply human-perceived good quality in scientific protocols. 
(4) The use of automatic metrics~\hyperref[LLMeval:1]{(i)} requires manual labor, which limits the transition to fully automatic evaluation.

    % \footnote{Includes metrics using statistical scoring. Examples can be found at~\hyperref[LLMeval:1]{(i)}.}

    % I want to imply that the high scores in traditional automatic metrics does not necessarily mean scientific protocols in good quality. 
    %% Traditional automatic metrics\footnote{Based on statistical evaluations. Further details~\hyperref[LLMeval:1]{here}.} frequently show a weak correlation with human evaluations, highlighting their inadequacy in precisely assessing the caliber of scientific protocols."

    % 예시 찾아서 보여주면 좋을듯? 
    
%프로토콜 생성의 한 단면만을 평가한다. 
% It evaluates only one aspect of protocol generation.

% 단편적으로 인식한다
%It perceives in a fragmented manner.

    % The use of traditional automatic metrics may not adequately reflect the quality or human-perceived readability of protocols.
    % 
% Traditional Metric사용하는 것이 충분한 반영을 하지 못한다 (이런 metric의 우위 does not necessarily mean semantic similarity) 

% (i) 실험을 통해서 supplementary에 빼기? 
% (ii) 예시 supplementary로 빼기? 

% 실험실에서 일어나는 액션은 제한되어 있다 > 슈도코드를 미리 정의해서 슈도코드 추출의 variation을 없앤다

% acc, precision, recall - ‘function 같은함수끼리만 argument 비교하게 된다‘

% (Add \textsc{Llam-eval} here?)

Here, we propose an evaluation framework that evaluates the capabilities of LLMs in SPFT: \textit{ProtoMed-LLM} (Figure~\ref{fig:diagram}). 
% Step 1

% 예시 supplementary ? 
% 실험할 때 shuffle 시도
% Action 업데이트 된 후에 업데이트 하기
% 각각 description 주어진 공간에 맞게 좀더 길게 ㄱㄱ (1줄내로) 

\begin{table*}[h] % The asterisk here makes the table span both columns
\centering
\small
\begin{tabular*}{\textwidth}{@{\extracolsep{\fill}}ll}
\hline
\textbf{Action Name} & \textbf{Description} \\
\hline
\verb|Transfer| & Move substances between containers using lab equipment, such as pipettes. \\
\verb|Centrifuge| & Spin at high speed to separate mixture components by density. \\
\verb|Vortex| & Mix solutions by creating a vortex for even distribution. \\
\verb|SetTemp| & Set specific temperatures for reactions or processes. \\
\verb|Wait| & Period of inactivity to allow reactions or condition stabilization. \\
\verb|Wash| & Rinse materials, often with solvents to remove contaminants. \\
\verb|Measure| & Quantify substances or properties using instruments. \\
\verb|Microscopy| & Use a microscope to observe and analyze cell morphology and structures. \\
\verb|CellDetachment| & Release adherent cells from a culture surface using enzymatic or mechanical methods. \\
\verb|CellCount| & Determine the number of cells in a sample using a hemocytometer or automated counter. \\
\verb|InvalidAction| & Undefined action due to documentation error or ambiguity. \\
\verb|OtherLanguage| & Text in non-English, indicating translation need. \\
\verb|NoAction| & Text not corresponding to any defined action. \\
\midrule
\verb|PCR| & Amplify DNA segments through Polymerase Chain Reaction. \\
\verb|Gel| & Separate molecules by size in a gel with electric field. \\
\verb|Culture| & Grow cells in lab to study behavior or for experimentation. \\
\verb|Dilute| & Reducing the concentration of a solution by adding solvent. \\
\hline
\end{tabular*}
\caption{\label{tab:actions}\textbf{Predefined Set of Actions.} List of actions performed in biological experiments and the corresponding descriptions. Actions above the line represent the basic actions, with the last three specifically designated for instances where a new protocol introduces an undefined action. Actions below represent the coarse-grained actions.}
%The details of coarse-grained actions can be found at Table~\ref{tab:course-grained-actions}. 
\end{table*}
First, we define a set of actions in advance (Table~\ref{tab:actions}), which eliminates individual action (pseudofunction) extraction step and variations of actions on each occasion. 
% Step 2
Second, we independently zero-shot prompted the target model and GPT-4~\citep{openai2023gpt4} to extract pseudocode from biology protocols, only using predefined actions as pseudofunctions. 
%(Figure~\ref{fig:one-shot-prompt-example})
% 수정필요
% % Step 3: LLam-Eval (G-Eval Method 에서 evaluator를 Llama-3로) 
Lastly, we use \textsc{Llam-Eval} to evaluate the response, treating the target model's pseudocode as a prediction ($\hat{y}$) and that of GPT-4's as a baseline ($y$).
\textsc{Llam-Eval} offers significant flexibility in terms of evaluation model, material, and criteria.
% Inspiration
This approach is inspired by the automated extraction of chemical synthesis actions from experimental procedures\footnote{A set of actions in chemistry labs were defined prior to the pseudocode extraction process.}~\citep{Vaucher2020}. 
% 의의
% \textsc{LLam-eval} Contribution
% 결과요약
We compared multiple LLMs to our framework, including GPT variations~\citep{brown2020language, openai2023gpt4}, Llama, Mixtral, Gemma, Cohere, and Gemini~\citep{geminiteam2024gemini}. 
% Cohere technical report 추가
% Dataset 설명은 intro에서 빼기 (bioplanner도 그랬음). 
% First, we collect protocols from \textit{protocol.io}~\citep{protocols.io}.
We find that GPT-4o and Cohere+ is a powerful scientific protocol formulator.

% 일부러 intro에서는 뺌
% This method is inspired my G-Eval~\citep{liu2023geval}. 

% This approach provides exceptional adaptability in the evaluation process.

% eval 대상이 되는 모델이 내놓는 output에 구애받지 않으며 parsing 과정에서 필요한 manual annotation 노력이 필요없게 된다. 

% (ADD \textsc{Llam-Eval} contribution and explaination)

We also introduce \textsc{Bioprot 2.0}, a larger dataset with scientific protocols and the corresponding pseudocodes that can aid LLMs in formulation and evaluation of SPFT.  

% previous = Bioprot 인데 이름 직접언급은 하지 않을 것

% We validated the efficacy of this dataset through real-world experiments by executing the protocols generated by the LLM, which were aided by \textsc{Bioprot 2.0}. 
% 우리 dataset 장점 추가
Overall, we make the following contributions: 
\begin{enumerate}\itemsep-3pt
    \item We propose \textit{ProtoMed-LLM}: a flexible, automatic framework for evaluating LLMs on SPFT using domain knowledge and \textsc{Llam-Eval}. 
    % 아래두개 합쳐서 위의 1개로 만듦
    % \item We implement domain knowledge in biology labs to evaluate LLMs on SPFT.
    % \item We introduce GPT as an evaluator to assess the capability of LLMs on SPFT.
    \item We propose \textsc{Llam-Eval}, an evaluation method that uses a form-filling paradigm offering significant flexibility in terms of evaluation model, material, and criteria.
    % 우리는 (next step prediction 같은) task정의해서 하는 evaluation에서 탈피한 LLM evaluation framework를 제안했다. 

    % by adopting G-Eval 
    % GPT Eval을 프로토콜에 최초로 도입함
    % contribution: 다른 분야에도 domain knowledge 도입하면 더 정확하게 eval 할수 있다. 

    % Bioprot 2.0 을 fine-tuning 해서 domain specific 한 scientific protocol generation에 이용할 수 있다.
    \item We introduce the \textsc{BioProt 2.0} dataset, featuring protocols and corresponding pseudocode for evaluating and aiding LLMs on SPFT. 

    % 우리의 입장에서는 bioplanner와 다르게 protocol 'generation' 이 맞는지 확인 필요
    % \item We have automatically generated and verified protocols in our biology lab.

\end{enumerate}

% Contribution 추가: API 안쓰고 (돈 안쓰고) 평가방식을 이용할 수 있다 
\section{Related Works}
% G-Eval Related Works에서 아이디어
% G-Eval의 related works > Task-specific Evaluators 에서는
% : Summarization tasks 위주 기술
% 특정 task 평가를 대상으로 하는 경우는 기존 문헌이 있다 / 하지만 특정 task가 SPFT인 경우는 거의없다 > bioplanner에서밖에 없다 논리 펼치기

% ------------ Storage ------------
% Prompt에 참고
% Dialogue Systems Can Generate Appropriate Responses without the Use of Question Marks? -- Investigation of the Effects of Question Marks on Dialogue Systems
% \citep{mizumoto2023dialogue}

% Limitation에 활용? 
% Tricking LLMs into Disobedience: Formalizing, Analyzing, and Detecting Jailbreaks
% \citep{rao2024tricking}

% 학회에서 본 논문 - 찾아도 안나옴
% Enhancing scientific document summariztion with research community perspective and background knowledge
% background knowledge for document summarization~\citep{}

% combination of {named-entity recognition (NER), question-answering (QA), relation extraction (RE)} in specific domain

% % 모델 종류- two proprietary models (GPT-3.5 and GPT-4), three open-source models (Mistral 7B, Falcon 40B, and Llama 2 70B)
% \citep{mortensen2024verbing}
% ---------------------------------

\noindent\textbf{Task-specific Evaluation} LLMs have been evaluated based on their performance in specific tasks.
% 다양한 LLMs 는 많은 경우에 specific한 task를 얼마나 잘 수행하였는지로 평가되어 왔다. 
% Task 1
Information extraction abilities were measured by the generated quality of summaries~\citep{ durmus-etal-2020-feqa, wang-etal-2020-asking}, paper reviews \citep{zhou-etal-2024-llm-reliable}, question correction~\citep{fan2024evaluating}, or combination of a few tasks~\cite{labrak2024zeroshot}.
% Task 2 - 추가하려고 했으나 필요없을듯
% 한계점: 하지만, 이러한 연구들의 specific task를 통해 comprehensive하지 못하고 굉장히 일부분만을 평가할 수 있게 된다.
However, these studies do not provide comprehensive evaluations and only assess very limited aspects, thus limiting their generalizability to other abilities or tasks.
\\

\noindent\textbf{LLM Evaluation on SPFT} Recent work proposes a three-step framework (Section~\ref{sec:introduction}) for the evaluation of scientific protocols in biology: \textit{BioPlanner} \citep{odonoghue-etal-2023-bioplanner}. This work evaluates GPT's performance in three tasks: next-step prediction, pseudocode generation, and pseudofunction retrieval. It employs statistical scoring methods including Levenshtein distance (\( \mathcal{L}_d \)) and BLEU ~\citep{BLEU} to measure the relevance between a baseline and generated protocols, despite their modest correlation with human judgments.\\

% GPT-4 평가하는것이 우위에 있다는 논리"which denounces us to the separation." 같은 어투? 

% Pesudofuctions in \textit{Bioplanner} are ought to be the corresponding actions in experiments. For instance, pesudofuction "\verb|def transfer_tissue|" and "\verb|def incubate_sample|" corresponds to transferring tissues to another tube and incubating samples for a given time.
% (insert Pseudocodes definition explanation) 

%G-Eval Related Works에서 아이디어
%\noindent\textbf{LLM-based Evaluators}
%\\
% 아래는 내 아이디어
% \noindent\textbf{(Evaluating) LLMs in Specific Domains}
\noindent\textbf{Domain-specific LLMs in Science} A Large number of LLMs have been trained, fine-tuned, or augmented for domain-specific uses. 
ChemBERTa/-2~\citep{chithrananda2020chemberta, ahmad2022chemberta2}, MatSciBERT~\citep{gupta2021matscibert}, MaterialsBERT~\citep{shetty2023general}, Chemcrow~\citep{bran2023chemcrow}, and LLM augmentation methods for various experiment-related tasks~\citep{guo2023large} has been introduced in chemistry. BioGPT~\citep{BioGPT}, BioBERT~\citep{BioBERT}, CamemBERT-bio~\citep{touchent2024camembertbio}, BlueBERT~\citep{peng2019transfer}, PubmedBERT~\citep{pubmedbert}, BioMegatron~\citep{shin2020biomegatron}, and ProtoCode~\citep{Jiang2024} has been introduced in biology. 
% 약간 관련없는 것 같아 임시 제외
% (PDDL) synthesize Python code~\citep{silver2023generalized}
% (위에 추가함) Bioplanner 인용한 논문 - PCR 함
% ProtoCode~\citep{Jiang2024}
\\
\\
\noindent\textbf{Evaluating LLMs with LLMs} Evaluation of LLMs encompasses a dual-method approach: 
% 뒤 내용에 쓰지 않은 내용 나중에 지우기
\begin{enumerate}[label=(\roman*)]\itemsep-3pt
    \item Statistical scoring: BLEU \citep{BLEU}, ROUGE \citep{lin-2004-rouge}, METEOR \citep{banerjee-lavie-2005-meteor}, Levenshtein Distance
    \label{LLMeval:1}
    \item Model-based scoring: G-Eval \citep{liu2023geval}, Prometheus \citep{kim2023prometheus}, BLEURT \citep{sellam2020bleurt}, Natural Language Inference (NLI)
    \label{LLMeval:2}
    \item Combination of \ref{LLMeval:1} and \ref{LLMeval:2}: GPTScore \citep{fu2023gptscore}, SelfCheckGPT \citep{manakul2023selfcheckgpt}, BERTScore \citep{zhang2020bertscore}, SciBERTScore \citep{odonoghue-etal-2023-bioplanner}, WMD~\citep{pmlr-v37-kusnerb15}, MoverScore \citep{zhao2019moverscore}, Question Answer Generation (QAG) Score 
    \label{LLMeval:3}
\end{enumerate}

% GPT-Score 한계점 추가
\noindent In tasks where reasoning is involved, \ref{LLMeval:2}\ref{LLMeval:3} outperforms \ref{LLMeval:1}. Previous work adopted \ref{LLMeval:1} with \ref{LLMeval:3} being minimal~\citep{odonoghue-etal-2023-bioplanner}. In this work, we adopt the notion of G-Eval~\citep{liu2023geval}, a framework for evaluating LLM-generated text, which prompts GPT with text and criteria, then scores based on its output.

\section{Methods}
% methods overview explanation 

\begin{figure*}[h!]
  \includegraphics[width=\textwidth]{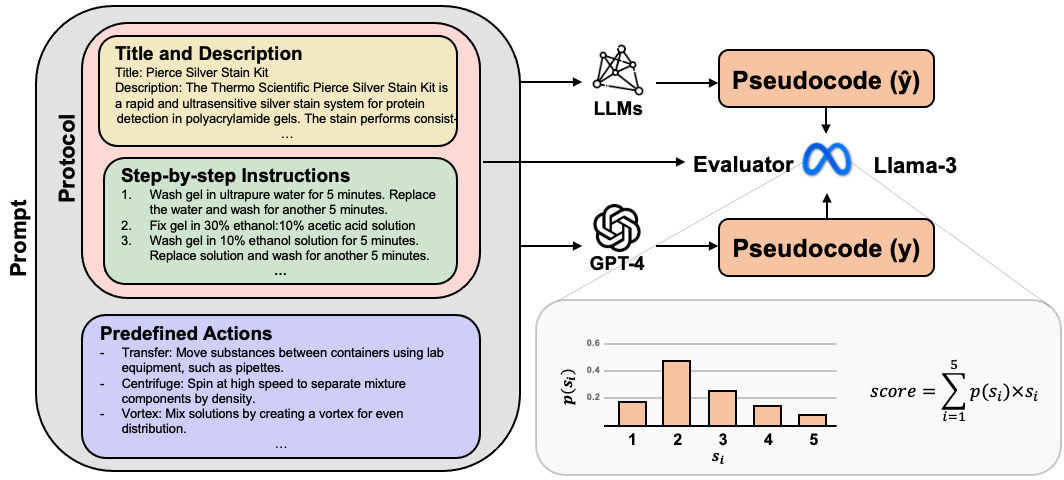}
  \caption{\textbf{The \textit{ProtoMed-LLM} Framework.}}
  \label{fig:diagram-large}
\end{figure*}

The \textit{ProtoMed-LLM} framework can evaluate the capability of LLMs on SPFT in three steps (Figure~\ref{fig:diagram-large}): (1) prompt the target LLM to generate pseudocode based on the given protocol, (2) repeat previous step for GPT-4, and (3) \textsc{Llam-Eval} for evaluation.
To utilize this framework, we curated protocols in biology (Section~\ref{sec:data-curation}), predefined actions performed in biology labs (Section~\ref{sec:defining-actions}), prompted LLMs for pseudocode generation (Section~\ref{sec:code-generation}), and prompted Llama-3 for evaluation (\textsc{Llam-Eval}) (Section~\ref{sec:llameval}). 

\subsection{Data Curation of Protocols in Biology}
\label{sec:data-curation}

% Dataset of Protocols in Biology
% protocol collection

% 구성 설명 추가? 고민.. : Each protocol includes a title, description, and step-by-step instructions. 
% 나중에 작성후 레퍼런스 잘 달렸는지 확인할 것

Each protocol is composed of three core elements: a title, description, and experimental steps. We curated the dataset through a process of collection and refinement. We collected a set of keywords relevant to biology. Then, we used a scoring system based on the number of keywords included in the description of each protocol from \textit{protocols.io}\footnote{A platform for reproducible protocol sharing provides access to more than 15k publicly available protocols, and has no limitations regarding the use of LLMs.}~\citep{protocols.io}. 
We refined the dataset collected in the previous step using automated and manual methods. (Appendix \ref{sec:appendix-DatasetCuration}.)

\subsection{Defining Actions}
\label{sec:defining-actions}

The defined actions are composed of two parts: (i) \textbf{basic actions} corresponding to a single action which can be performed directly in biology labs, and (ii) \textbf{coarse-grained actions} which corresponds to a large set of basic actions repeated throughout various protocols. Defined actions were reviewed by experts with intensive experiences in biology experiments. The target model specifies the arguments for each action on each occasion.\\

% 'Arguments'가 무엇인지 정의 작성? 

\noindent\textbf{Basic Actions} Since the repertoire of actions executed in biology labs is confined to a finite set of actions, we defined a set of actions performed in biology labs prior to the extraction of pseudocode from protocols (Table~\ref{tab:actions}). We performed a comprehensive literature review to define the set of basic actions performed in biology labs. \\

% 다른 representations로 인한 performance degradation 이야기도 포함시키는 것이 맞는데, 근거가 없어서 포함시키는 것이 맞는지는 모르겠다. 

% We have heuristically reviewed biology textbooks, papers, and experimental protocols.

% Action이 무엇인지에 대한 설명

\noindent\textbf{Coarse-grained Actions} 
We observed that a series of complex, repetitive actions can be effectively encapsulated and described by a single, comprehensive action. For instance, the process of diluting a solution is conceptually straightforward and can possibly defined by basic actions. However, this involves intricate calculations and logical reasoning, which can result in performance degradation by calculation mistakes and posing variations in representations of an identical process. To this end, we coarse-grained these complex set of actions into a singular action. 
%(details in Appendix~\ref{sec:course-grained}). 

% For example, dilution of a solution may simply be defined as an action by only using basic actions, but involves complicated calculation and logical reasoning, which can lead to performance degradation.
% For example, Polymerase Chain Reaction (PCR) can be explained through about twenty predefined actions. 

% Argument는 protocol의 핵심 요소인 통제변인(?)을 추출하는 능력을 정성적으로 포함하고 있으므로 평가 대상에 포함된다(?)

% 어떻게 action 병합했는지 예시포함 supplementary에 적기

% ---- (여기서부터) ---- 
\subsection{Prompting Pseudocode Generation}
\label{sec:code-generation}

% Prompt 관련내용 추가
% prompt 어떤식으로 했는지, 등등 detail 적기
To evaluate the target LLMs on SPFT, we prompted the models to generate pseudocode based on a protocol collected at Section~\ref{sec:data-curation}. Models are instructed to use only the actions defined in Section~\ref{sec:defining-actions} as the function name. However, they were allowed to define the arguments for each pseudofunction as needed for each occasion.
%If applicable (가능한경우), 고정되어있는 prompt인 instruction과 predefined actions는 system message 에, protocol은 user message에 제공하였다. 
If applicable, the fixed prompt, including the instructions and predefined actions, was provided in the system message, while the protocol was included in the user message.
% 구체적인 모델명 추가
In this work, we prompted GPT-3~\citep{brown2020language}, GPT-4~\citep{openai2023gpt4}, Gemini~\citep{geminiteam2024gemini}, Claude3~\citep{Anthropic2023Claude3}, and Cohere. 
%The prompt can be found at Appendix~\ref{app:pseudocode-generation}. 
Below is the prompt for generating pseudocode based on the given protocol. Note that actions and corresponding descriptions presented in Table~\ref{tab:actions} are placed at \textit{\{actions\}}.
\\
\\
\textit{You are an AI that generates Python pseudocode for biology protocols. This pseudocode must accurately describe a complete scientific protocol to obtain a result. You will be provided with the title, description, and steps of the biology protocol, and your task is to convert it to Python pseudocode.}
\\
\\
\textit{You may define the arguments on your own. You must ONLY use these functions.}

\textit{\{actions\}}
\\
\\
\textit{Do NOT provide any captions. ONLY present the pseudocode and pseudofunctions used inside the code. Present the pseudofunctions at the beginning and then the pseudocode. Do NOT provide any descriptions inside the code.
}
\\
\\
\textit{title: \{title\}}

\noindent\textit{description: \{description\}}

\noindent\textit{steps:} \textit{\{steps\}}

\subsection{Metrics and Evaluation}
\label{sec:metrics-eval}
We observe that using automatic metrics~\hyperref[LLMeval:1]{(i)} necessitates manual annotation of functions and pseudocodes each time, which significantly hampers the automation of the evaluation process. Moreover, evaluating the function and input\footnote{Input refers to the function parameters and arguments.} separately falls short of flexible and comprehensive evaluation in a protocol manner.
% (ADD flexible meaning?)

To this end, we propose \textsc{Llam-Eval}, an automatic, flexible prompt-based framework to evaluate the quality of LLM responses. This framework requires three elements: two input texts (one serving as the baseline and the other as the target) and an evaluator LLM: Llama-3\footnote{Llama3-70b}. 
% Llam-Eval 설명
This method encompasses predefining a set of scores\footnote{$s_1$=1 and $s_n$=5 is set in this work.} $S=\{s_1, s_2, ..., s_n\}$, prompting Llama-3 to rate the outcomes of a target LLM with that of GPT-4 in the scale of $S$, calculating the probability of each score $p(s_{i})$, and calculating the final score as following. This method is inspired by G-Eval~\citep{liu2023geval}.

\[
\text{score} = \sum_{i=1}^{n} s_{i}p(s_{i})
\] 
   
Llama-3 is prompted to evaluate according to one criterion at a time. The original prompts targeting summarizing tasks are modified to perform evaluation on SPFT.  
% 2가지 새로운 기준 
% Coverage: The extent to which the text addresses all aspects of the topic.
% Precision: The exactness and accuracy of the expressions and terminology used in the text.
In this work, we evaluate the pseudocode generated by the target LLM based on six criteria: the four original criteria used in G-Eval~\citep{liu2023geval} (\verb|Coherence|, \verb|Consistency|, \verb|Fluency|, and \verb|Relevance|) and two criteria we propose (\verb|Precision|, and \verb|Coverage|), considering the context of SPFT. For example, the definition of \verb|Coherence| is: 
\\\\
\textit{Coherence (1-5) - the overall quality of all lines in the pseudocode. The target pseudocode should not be a rough overview but should provide a precise description of a baseline pseudocode.}
\\\\
The definitions of other criteria in prompts can be found at Appendix~\ref{app:Metrics-Evaluation}. 
To automatically implement chain-of-thoughts (CoT) in the evaluation process, we instructed GPT-4 to create specific evaluation steps for each criterion. GPT is capable of producing these evaluation steps by itself~\citep{liu2023geval}. GPT-4 was given a task and evaluation criteria, then prompted to generate the evaluation steps using a form-filling paradigm. An example prompt containing GPT-4 generated instructions for evaluation can be found at Appendix~\ref{app:Metrics-Evaluation}.
% For evaluation tasks, some criteria need a more detailed
% evaluation instruction beyond the simple definition,
% and it is time-consuming to manually design such
% evaluation steps for each task. We find that LLM
% can generate such evaluation steps by itself. The
% CoT can provide more context and guidance for the
% LLM to evaluate the generated text, and can also
% help to explain the evaluation process and results.
% For example, for evaluating coherence in text summarization, we add a line of “Evaluation Steps:” to
% the prompt and let LLM to generate the following
% CoT automatically:
We also implemented an automatic feedback loop to regenerate the response up to five or ten times if the output did not contain scores. We evaluated using two baselines: the GPT-generated pseudocode and the original protocol. 

This approach is not constrained by the output structure of the target models, eliminates the need for manual annotation efforts during the parsing process as required in reference-based metrics, enables a comprehensive evaluation, and thereby makes \textit{ProtoMed-LLM} significantly more flexible and automatic.

% ~\citep{liu2023chatgptpowered}
% Formally, given a set of
% scores (like from 1 to 5) predefined in the prompt
% S = {s1, s2, ..., sn}, the probability of each score
% p(si) is calculated by the LLM, and the final score
% is:
% \textbf{Baseline} --> G-Eval = reference free metric

% 쓸 때는 비교하기 위해서 썼다기 보다는 원본 논문같이 chain-of-thought 어쩌고
% Bioplanner랑 똑같은 방법 사용
To ensure compatibility, we also use conventional reference-based metrics: Normalized Levenshtein distance (\( \mathcal{L}_{dn}\)) for function names,  BLEU~\citep{BLEU}, precision, recall, and SciBERTScore~\citep{odonoghue-etal-2023-bioplanner} for function inputs. % Detailed definitions are at Appendix~\ref{app:Metrics-Evaluation}.
% next step prediction 없어져서 아래 metric은 못쓸수도 있다
SciBERTScore is calculated using the encoded \textbf{pred}icted $\mathcal{E}(a_{i}^{\text{pred}})$ and \textbf{b}ase\textbf{l}ine values  $\mathcal{E}(a_{i}^{\text{BL}})$
using the SciBERT~\citep{beltagy-etal-2019-scibert} sentence encoder $\mathcal{E}$.
 \[\text{SciBERTScore}=\frac{1}{N} \sum_{i=0}^{N} \frac{\langle \mathcal{E}(a_{i}^{\text{pred}}), \mathcal{E}(a_{i}^{\text{BL}}) \rangle}{\| \mathcal{E}(a_{i}^{\text{pred}}) \| \| \mathcal{E}(a_{i}^{\text{BL}}) \|}
\]
\subsection{Evaluator LLM Selection}
\label{sec:eval-selection}
% self-self 비교해서 가장 5점에 가깝게 나온 모델을 baseline으로 삼기
To select a specific LLM as an evaluator, we propose \textit{self-self comparison task} as a baseline, where an LLM generates a pseudocode\footnote{Pseudocode with pseudofunctions defined at the beginning to be precise.} for a protocol and then evaluates the score using the same LLM against the generated pseudocode. For example, this means evaluating GPT-4 generated pseudocode against the same pseudocode using GPT-4. 
Our assumption was that the 
score should be close to the maximum\footnote{maximum score $s_n=5$ in this work} when the baseline and target pseudocode are the same. Our goal was to select the model with the best results as the evaluator. We evaluated each model based on six criteria in Section~\ref{sec:metrics-eval}. More details in Appendix~\ref{app:eval-llm-selection}. While using \textsc{G-Eval}, 
% to assess \verb|consistency|
we encountered instances where the output was a sentence instead of a score (number). To address this issue, we modified the parameters, dataset, and prompts. Further details are in Appendix~\ref{app:implementation-details}.

\subsection{Evaluating LLMs using \textsc{Llam-Eval}}
\label{sec:llameval}

Using \textsc{Llam-Eval}, we evaluate across three tasks for each model: (1) GPT-4 generated pseudocode as a baseline with predefined actions given in prompt, (2) the same task with no predefined actions, (3) the original protocol as a baseline with predefined actions. We evaluate GPT variations~\citep{brown2020language, openai2023gpt4}, Llama, Mixtral, Gemma, Cohere, and Gemini~\citep{geminiteam2024gemini}.
Details are in Appendix~\ref{app:eval-llms-spft}.
% of model versions 
 
\subsection{Implementation Details} 

 To ensure a fair evaluation of LLMs, we considered additional factors that may affect performance and present several settings. We consider that LLMs tend to perform better when the actions are presented in the same order as in the protocol. While previous work extracted different actions from each protocol\footnote{In previous work, this required shuffling, as LLMs presented the pseudofunctions in the same order as in the protocol.}, we predefined the actions which is equivalent to shuffling.

\begin{table*}[!t]
\centering
%\footnotesize % Reduce the font size to 'footnotesize'
\setlength{\tabcolsep}{4pt} % Reduce space between columns
\begin{tabular*}{\textwidth}{@{\extracolsep{\fill}}l|cccc|cc}
\toprule
 & \multicolumn{4}{c|}{Original Criteria} & \multicolumn{2}{c}{New Criteria} \\ 
Models & Coherence & Consistency & Fluency & Relevance & Precision & Coverage \\ 
\midrule
GPT-4o & $4.95 \pm 0.26$ & $4.98 \pm 0.25$ & $4.95 \pm 0.27$ & $4.93 \pm 0.44$ & $4.97 \pm 0.23$ & $4.95 \pm 0.08$ \\
GPT-4 & $4.98 \pm 0.23$ & $4.99 \pm 0.19$ & $4.99 \pm 0.19$ & $4.99 \pm 0.14$ & $4.99 \pm 0.18$ & $ 4.99\pm 0.17$ \\
GPT-3.5 & $4.96 \pm 0.23$ & $4.97 \pm 0.21$ & $4.77 \pm 0.52$ & $4.96 \pm 0.25$ & $4.95 \pm 0.30$ & $4.99 \pm 0.12$ \\
\textbf{Llama-3} & $\textbf{5.00} \pm 0.02$ & $\textbf{5.00} \pm 0.00$ & $\textbf{5.00} \pm 0.00$ & $\textbf{5.00} \pm 0.00$ & $\textbf{5.00} \pm 0.00$ & $\textbf{5.00} \pm 0.06$ \\
\bottomrule
\end{tabular*}
\caption{\textbf{\textit{Self-Self Comparison Task} Results}: We report the mean and standard deviation of scores over five or ten runs. Values in bold indicate the highest scores for each criterion. Higher values for all metrics represent better performance. Note that a larger dataset was used for this task. Details in Appendix~\ref{app:eval-llm-selection}. 
}
\label{tab:evaluator-comparison}
\end{table*}

\section{Analysis}

\subsection{Evaluator LLM Selection} 
Llama-3 achieved the highest scores across all six tasks, while there were small differences across models (Table~\ref{tab:evaluator-comparison}). We chose Llama-3 as an evaluator, which is free of cost to date. Note that evaluations for other models not presented in the table were not feasible, as numerical responses were not generated. More details are in Appendix~\ref{app:eval-llm-selection}.

% However, we observed that 다른 모델들과 조금씩의 차이만이 있다는 것을 알 수 있었다. 따라서 evaluator로 가장 저렴한 3.5를 선택했다. Note that other model에서는 response에서 숫자가 나오지 않아 evalution자체가 성립되지 않았다.

% fragments
% fluency may be not quite right for our situation
% reason for using this method is that

\subsection{Evaluating LLMs on SPFT}

% \noindent\textbf{GPT-4o and Cohere+ is a powerful protocol formulator}
% GPT-generated pseudocode as a baseline
Our results show that GPT-4o and Cohere+ is a powerful protocol formulator (Table~\ref{tab:results-noaction}).
% which is compatible to previous work (bioplanner) 
We found our work compatible to previous work~\citep{odonoghue-etal-2023-bioplanner}.
% (ADD efficacy of out methods) Our results using traditional metrics  (Table~\ref{tab:results})
% Our method offers the advantage of utilizing any manually annotated results as a baseline for comparison.
\begin{table*}[htbp!]
\footnotesize % Reduce the font size to 'footnotesize'
\centering
\setlength{\tabcolsep}{4pt} % Reduce space between columns
\begin{tabular*}{\textwidth}{@{\extracolsep{\fill}}l|cc|cccc|cc|c}
\toprule
 & \multicolumn{2}{c|}{Prompt} & \multicolumn{4}{c|}{Original Criteria} & \multicolumn{2}{c|}{New Criteria} &  \\
Models & Ac & Pr & Coherence & Consistency & Fluency & Relevance & Precision & Coverage & Average\\ 
\midrule
GPT-4o & \ding{51} & \ding{55} & $\textbf{4.10} \pm 0.79$ & $\textbf{3.80} \pm 0.85$ & $\textbf{3.86} \pm 0.67$ & $\textbf{4.32} \pm 0.71$ & $\textbf{4.02} \pm 0.65$ & $\textbf{4.26} \pm 0.73$ & $\textbf{4.06}$  \\
       & \ding{55} & \ding{55} & $\textbf{4.28} \pm 0.50$ & $\textbf{3.94} \pm 0.64$ & $\textbf{4.04} \pm 0.37$ & $\textbf{4.45} \pm 0.54$ & $\textbf{4.18} \pm 0.41$ & $\textbf{4.39} \pm 0.50$ & $\textbf{4.21}$  \\
       & \ding{51} & \ding{51} & $\textbf{4.29} \pm 0.57$ & $\underline{4.73} \pm 0.50$ & $4.42 \pm 0.53$ & $\textbf{4.75} \pm 0.48$ & $\textbf{3.90} \pm 0.48$ & $\textbf{4.67} \pm 0.56$ & $\underline{4.46}$  \\
\rowcolor{LightGray} GPT-4 & \ding{51} & \ding{55} & $5.00 \pm 0.00$ & $5.00 \pm 0.00$ & $5.00 \pm 0.08$ & $5.00 \pm 0.00$ & $4.99 \pm 0.11$ & $5.00 \pm 0.00$ & $5.00$  \\
\rowcolor{LightGray} (Baseline) & \ding{55} & \ding{55} & $5.00 \pm 0.00$ & $5.00 \pm 0.00$ & $5.00 \pm 0.04$ & $5.00 \pm 0.03$ & $5.00 \pm 0.03$ & $5.00 \pm 0.00$ & $5.00$  \\
\rowcolor{LightGray} & \ding{51} & \ding{51} & $4.32 \pm 0.53$ & $4.70 \pm 0.58$ & $4.53 \pm 0.51$ & $4.75 \pm 0.44$ & $3.99 \pm 0.29$ & $4.67 \pm 0.48$ & $4.49$  \\
GPT-3.5 & \ding{51} & \ding{55} & $3.61 \pm 0.97$ & $3.51 \pm 1.02$ & $3.58 \pm 0.85$ & $4.11 \pm 0.78$ & $3.82 \pm 0.73$ & $3.90 \pm 0.83$ & $3.75$  \\
        & \ding{55} & \ding{55} & $3.83 \pm 0.82$ & $3.71 \pm 0.81$ & $3.76 \pm 0.68$ & $4.19 \pm 0.64$ & $3.96 \pm 0.57$ & $3.97 \pm 0.71$ & $3.90$  \\
        & \ding{51} & \ding{51} & $4.13 \pm 0.65$ & $\textbf{4.76} \pm 0.49$ & $\underline{4.48} \pm 0.52$ & $\underline{4.69} \pm 0.49$ & $3.79 \pm 0.58$ & $\underline{4.49} \pm 0.67$ & $4.39$  \\
\midrule
Llama3-8b & \ding{51} & \ding{55} & $2.25 \pm 1.00$ & $1.93 \pm 0.99$ & $2.27 \pm 0.83$ & $2.39 \pm 1.08$ & $2.61 \pm 0.96$ & $2.56 \pm 1.09$ & $2.33$  \\
          & \ding{55} & \ding{55} & $2.90 \pm 0.89$ & $2.69 \pm 0.92$ & $3.02 \pm 0.88$ & $3.41 \pm 0.81$ & $3.47 \pm 0.70$ & $3.19 \pm 0.82$ & $3.12$  \\
          & \ding{51} & \ding{51} & $2.80 \pm 1.02$ & $3.00 \pm 1.26$ & $3.10 \pm 1.00$ & $3.39 \pm 1.09$ & $2.93 \pm 0.92$ & $3.27 \pm 1.06$ & $3.08$  \\
Llama3-70b & \ding{51} & \ding{55} & $3.61 \pm 0.94$ & $3.14 \pm 1.10$ & $3.53 \pm 0.82$ & $3.73 \pm 0.97$ & $3.72 \pm 0.70$ & $3.77 \pm 0.79$ & $3.58$  \\
           & \ding{55} & \ding{55} & $\underline{3.98} \pm 0.64$ & $\underline{3.72} \pm 0.75$ & $3.92 \pm 0.49$ & $\underline{4.20}\pm 0.57$ & $\underline{4.03} \pm 0.36$ & $\underline{4.09} \pm 0.53$ & $\underline{3.99}$  \\
           & \ding{51} & \ding{51} & $4.02 \pm 0.75$ & $4.17 \pm 0.98$ & $4.15 \pm 0.66$ & $4.37 \pm 0.74$ & $3.78 \pm 0.59$ & $4.25 \pm 0.69$ & $4.12$  \\
\midrule
Mixtral & \ding{51} & \ding{55} & $3.41 \pm 1.03$ & $2.90 \pm 1.13$ & $3.57 \pm 0.83$ & $3.36 \pm 1.14$ & $3.68 \pm 0.77$ & $3.54 \pm 0.93$ & $3.41$  \\
        & \ding{55} & \ding{55} & $3.95 \pm 0.68$ & $3.68 \pm 0.79$ & $3.94 \pm 0.53$ & $4.18 \pm 0.66$ & $4.05 \pm 0.43$ & $4.00 \pm 0.61$ & $3.97$  \\
        & \ding{51} & \ding{51} & $4.06 \pm 0.69$ & $4.32 \pm 0.84$ & $4.28 \pm 0.59$ & $4.37 \pm 0.71$ & $3.88 \pm 0.44$ & $4.31 \pm 0.70$ & $4.21$  \\
\midrule
Gemma-7b & \ding{51} & \ding{55} & $3.06 \pm 0.97$ & $2.81 \pm 1.03$ & $3.47 \pm 0.86$ & $3.52 \pm 0.93$ & $3.55 \pm 0.78$ & $3.19 \pm 0.89$ & $3.27$  \\
         & \ding{55} & \ding{55} & $2.93 \pm 0.85$ & $2.66 \pm 0.88

$ & $3.63 \pm 0.76$ & $3.61 \pm 0.71$ & $3.66 \pm 0.61$ & $3.06 \pm 0.80$ & $3.26$  \\
         & \ding{51} & \ding{51} & $3.81 \pm 0.75$ & $4.13 \pm 0.83$ & $4.25 \pm 0.61$ & $4.26 \pm 0.75$ & $3.76 \pm 0.60$ & $3.94 \pm 0.79$ & $4.02$  \\
\midrule
Cohere+ & \ding{51} & \ding{55} & $\underline{3.95} \pm 0.74$ & $\underline{3.63} \pm 0.87$ & $\underline{3.87} \pm 0.60$ & $\underline{4.11} \pm 0.74$ & $\underline{3.98} \pm 0.50$ & $\underline{4.07} \pm 0.63$ & $\underline{3.94}$  \\
        & \ding{55} & \ding{55} & $3.97 \pm 0.60$ & $3.71 \pm 0.73$ & $\underline{3.95} \pm 0.46$ & $4.15 \pm 0.56$ & $\underline{4.03} \pm 0.38$ & $4.04 \pm 0.50$ & $3.98$  \\
        & \ding{51} & \ding{51} & $4.44 \pm 0.52$ & $4.63 \pm 0.61$ & $\textbf{4.53} \pm 0.52$ & $4.73 \pm 0.47$ & $4.04 \pm 0.30$ & $4.66 \pm 0.49$ & $\textbf{4.50}$  \\
Cohere & \ding{51} & \ding{55} & $3.51 \pm 0.91$ & $3.06 \pm 1.02$ & $3.56 \pm 0.74$ & $3.66 \pm 0.87$ & $3.71 \pm 0.63$ & $3.70 \pm 0.76$ & $3.53$  \\
       & \ding{55} & \ding{55} & $3.71 \pm 0.68$ & $3.44 \pm 0.83$ & $3.83 \pm 0.56$ & $4.05 \pm 0.53$ & $3.94 \pm 0.41$ & $3.84 \pm 0.56$ & $3.80$  \\
       & \ding{51} & \ding{51} & $\underline{3.98} \pm 0.63$ & $4.11 \pm 0.87$ & $4.14 \pm 0.51$ & $4.29 \pm 0.63$ & $3.83 \pm 0.48$ & $4.24 \pm 0.64$ & $4.10$  \\
\midrule
Gemini-1.0 & \ding{51} & \ding{55} & $2.77 \pm 1.09$ & $2.30 \pm 1.08$ & $2.90 \pm 0.95$ & $2.80 \pm 1.10$ & $3.13 \pm 0.92$ & $3.15 \pm 1.01$ & $2.84$  \\
& \ding{55} & \ding{55} & $3.46 \pm 0.93$ & $3.22 \pm 1.01$ & $3.59 \pm 0.83$ & $3.89 \pm 0.77$ & $3.80 \pm 0.69$ & $3.66 \pm 0.79$ & $3.60$  \\
& \ding{51} & \ding{51} & $3.37 \pm 0.93$ & $3.68 \pm 1.11$ & $3.73 \pm 0.80$ & $3.87 \pm 0.87$ & $3.42 \pm 0.78$ & $3.86 \pm 0.84$ & $3.66$  \\
Gemini-2.0 & \ding{51} & \ding{55} & $3.09 \pm 1.05$ & $2.53 \pm 1.10$ & $3.75 \pm 0.70$ & $2.98 \pm 1.08$ & $3.63 \pm 0.73$ & $3.43 \pm 0.89$ & $3.24$  \\
& \ding{55} & \ding{55} & $3.88 \pm 0.82$ & $3.61 \pm 0.91$ & $4.11 \pm 0.60$ & $4.13 \pm 0.73$ & $4.14 \pm 0.54$ & $3.93 \pm 0.73$ & $3.97$  \\
& \ding{51} & \ding{51} & $3.80 \pm 0.80$ & $3.95 \pm 0.97$ & $4.30 \pm 0.58$ & $4.18 \pm 0.72$ & $3.80 \pm 0.49$ & $4.11 \pm 0.68$ & $4.02$  \\
Gemini-1.5 & \ding{51} & \ding{55} & $3.02 \pm 1.05$ & $2.48 \pm 1.02$ & $3.10 \pm 0.93$ & $2.97 \pm 1.07$ & $3.32 \pm 0.84$ & $3.42 \pm 0.93$ & $3.05$  \\
& \ding{55} & \ding{55} & $4.12 \pm 0.66$ & $3.86 \pm 0.72$ & $4.03 \pm 0.55$ & $4.33 \pm 0.62$ & $4.13 \pm 0.50$ & $4.21 \pm 0.59$ & $4.11$  \\
& \ding{51} & \ding{51} & $3.34 \pm 0.95$ & $3.62 \pm 1.04$ & $3.76 \pm 0.76$ & $3.81 \pm 0.86$ & $3.36 \pm 0.77$ & $3.80 \pm 0.84$ & $3.61$  \\
\bottomrule
\end{tabular*}
\caption{\textbf{\textit{ProtocoLLM} Evaluation Results} of three tasks for each model: (1) GPT-4 generated pseudocode as a baseline with predefined actions given in prompt, (2) the same task with no predefined actions, (3) the original protocol as a baseline with predefined actions. '\textbf{Ac}' and '\textbf{Pr}' represent whether the predefined \textbf{ac}tions and the original \textbf{pr}otocol were given for evaluation, respectively.  We report the mean, standard deviation, and average of scores over five runs. The best and second best performance besides a baseline (GPT-4) for each criterion and task is bolded and underlined, respectively. The scores range from a minimum of 1 to a maximum of 5. Higher values for all metrics represent better performance.}
\label{tab:results-noaction}
\end{table*}
%Table~\ref{tab:results-noaction}
%\noindent We report evaluation results in Table~\ref{tab:results}.
\begin{table*}[htbp!]
\centering
\footnotesize % Reduce the font size to 'footnotesize'
\setlength{\tabcolsep}{4pt} % Reduce space between columns
\begin{tabular*}{\textwidth}{@{\extracolsep{\fill}}l|c|ccccccc}
\toprule
Models & Actions & Precision & Recall & SciBERT & BLEU & \( \mathcal{L}_{dn}\) \\ 
\midrule
GPT-4o & \ding{51} & $0.581 \pm 0.390$ & $0.548 \pm 0.414$ & $0.783 \pm 0.111$ & $0.102 \pm 0.189$ & \underline{$0.216 \pm 0.110$} \\
 & \ding{55} & $0.600 \pm 0.375$ & $0.620 \pm 0.373$ & $\textbf{0.778} \pm 0.103$ & $0.118 \pm 0.188$ & $0.214 \pm 0.106$ \\
\rowcolor{LightGray} GPT-4 & \ding{51} & $1.00 \pm 0.00$ & $1.00 \pm 0.00$ & $1.00 \pm 0.00$ & $0.905 \pm 0.198$ & $0.055 \pm 0.129$ \\
\rowcolor{LightGray} (baseline) & \ding{55} & $1.00 \pm 0.00$ & $1.00 \pm 0.00$ & $1.00 \pm 0.00$ & $0.911 \pm 0.173$ & $0.021 \pm 0.043$ \\
GPT-3.5 & \ding{51} & $0.817 \pm 0.308$ & $0.425 \pm 0.404$ & $0.766 \pm 0.115$ & $0.102 \pm 0.205$ & $\textbf{0.205} \pm 0.117$ \\
 & \ding{55} & $0.732 \pm 0.357$ & $0.572 \pm 0.378$ & $0.742 \pm 0.099$ & $0.099 \pm 0.178$ & $\textbf{0.200} \pm 0.106$ \\
\midrule
Llama3-8b & \ding{51} & $0.763 \pm 0.323$ & $0.708 \pm 0.411$ & $0.801 \pm 0.128$ & $0.135 \pm 0.329$ & $0.413 \pm 0.351$ \\
 & \ding{55} & $0.759 \pm 0.322$ & $0.570 \pm 0.352$ & $0.744 \pm 0.100$ & $0.075 \pm 0.174$ & $0.242 \pm 0.133$ \\
Llama3-70b & \ding{51} & $0.825 \pm 0.319$ & $\textbf{0.917} \pm 0.220$ & $\textbf{0.883} \pm 0.136$ & $\textbf{0.563} \pm 0.464$ & $0.287 \pm 0.203$ \\
 & \ding{55} & $0.812 \pm 0.268$ & $\textbf{0.769} \pm 0.260$ & $0.772 \pm 0.097$ & $\textbf{0.161} \pm 0.210$ & $0.206 \pm 0.095$ \\
\midrule
Mixtral & \ding{51} & $0.855 \pm 0.280$ & $0.605 \pm 0.393$ & $0.784 \pm 0.120$ & $0.135 \pm 0.288$ & $0.603 \pm 0.366$ \\
 & \ding{55} & $0.754 \pm 0.291$ & $\underline{0.735} \pm 0.290$ & $0.771 \pm 0.093$ & $0.130 \pm 0.215$ & $0.499 \pm 0.261$ \\
\midrule
Gemma-7b & \ding{51} & $\underline{0.911} \pm 0.249$ & $0.641 \pm 0.406$ & $0.838 \pm 0.139$ & $0.205 \pm 0.342$ & $0.243 \pm 0.130$ \\
 & \ding{55} & $\textbf{0.849} \pm 0.261$ & $0.651 \pm 0.337$ & $\underline{0.775} \pm 0.116$ & $0.092 \pm 0.180$ & $0.221 \pm 0.096$ \\
\midrule
Cohere+ & \ding{51} & $0.646 \pm 0.373$ & $0.548 \pm 0.352$ & $0.767 \pm 0.110$ & $0.075 \pm 0.172$ & $0.363 \pm 0.300$ \\
 & \ding{55} & $0.600 \pm 0.366$ & $0.604 \pm 0.368$ & $0.744 \pm 0.100$ & $0.095 \pm 0.153$ & $0.325 \pm 0.265$ \\
Cohere & \ding{51} & $0.645 \pm 0.361$ & $0.551 \pm 0.380$ & $0.717 \pm 0.097$ & $0.077 \pm 0.193$ & $0.360 \pm 0.247$ \\
 & \ding{55} & $\underline{0.767} \pm 0.295$ & $0.630 \pm 0.314$ & $0.750 \pm 0.099$ & $0.091 \pm 0.165$ & $\underline{0.204} \pm 0.105$ \\
\midrule
Gemini-1.0 & \ding{51} & $0.852 \pm 0.319$ & $0.867 \pm 0.313$ & $\underline{0.875} \pm 0.133$ & $\underline{0.444} \pm 0.497$ & $0.410 \pm 0.699$ \\
 & \ding{55} & $0.758 \pm 0.319$ & $0.584 \pm 0.360$ & $0.765 \pm 0.111$ & $0.112 \pm 0.211$ & $0.247 \pm 0.182$ \\
Gemini-2.0 & \ding{51} & $\textbf{0.942} \pm 0.147$ & $0.878 \pm 0.288$ & $0.843 \pm 0.165$ & $0.342 \pm 0.415$ & $0.381 \pm 0.254$ \\
 & \ding{55} & $0.736 \pm 0.350$ & $0.651 \pm 0.339$ & $0.758 \pm 0.104$ & $0.128 \pm 0.197$ & $0.308 \pm 0.268$ \\
Gemini-1.5 & \ding{51} & $0.889 \pm 0.258$ & $\underline{0.896} \pm 0.202$ & $0.814 \pm 0.163$ & $0.355 \pm 0.461$ & $0.371 \pm 0.217$ \\
 & \ding{55} & $0.628 \pm 0.377$ & $0.682 \pm 0.367$ & $0.773 \pm 0.101$ & $\underline{0.135} \pm 0.205$ & $0.214 \pm 0.116$ \\

\bottomrule
\end{tabular*}
\caption{\textbf{Evaluation Results Using Reference-Based Metrics.} Comparison with and without predefined actions given in prompts.
% \textbf{Protocol Generation Evaluation Results} of models with and without predefined actions.
We report mean and standard deviation of scores. The best and second best performance for each criterion is bolded and underlined, respectively. Except for \( \mathcal{L}_{dn}\), higher values for all metrics represent better performance.}
\label{tab:results}
\end{table*}
% 모델별로 강점이 있는 Criteria가 있을 수 있다) 
%This conveys the following: 
%Each model can exhibit specific strengths in different criteria. Llama-3 maybe not a good formulator but an evaluator.
% (IMPORTANT add also add to abstract?, intro, conclusion)
%(ADD \textsc{Llam-Eval} makes it flexibile in terms of eval criteria, input type, scoring scale)
\\\\
\noindent\textbf{Is applying domain knowledge an effective strategy for evaluation?}
We applied domain knowledge by predefining the finite set of actions performed in biology labs. 
To evaluate the efficacy of this method, we compare the responses generated with predefined actions included in the prompts to those generated without them (Table~\ref{tab:results}). The performance is enhanced for most models, with the exception of the \textit{Recall}. Further research should be conducted to explore these findings.
\\\\
\noindent\textbf{Can the original protocol itself serve as a baseline?}
Evaluation of LLMs in SPFT in previous work requires manual processes and pseudocode extraction step in SPFT. 
% 추가
% 한발 더 나아가 오리지날을 baseline로 사용할 수 있으면 가능? 획기적
However, evaluation using the original protocol itself completely eliminates the manual processes of pseudofunction evaluation and the GPT-generated pseudocode extraction step, thereby enhancing flexibility and automation.
%This enables resource-efficient evaluation on SPFT. 
To this end, we evaluate using the original protocol as a baseline.

While scores obtained using this approach is not close to the maximum score (Table~\ref{tab:results-noaction}), we observe that the relative ranking of the models remains relevant to the results of using the pseudocode as a baseline.
% 숫자가 너무 비슷해서 아래 제외
% except for \verb|Consistency| and \verb|Fluency|.
% We propose two potential reasons for this phenomenon: (1) Evaluator Llama-3 cannot precised evaluate these two criteria or (2) the generated pseudocode itself has a relative to .
% preliminary study
% This indicates that this method can fully exploit the advantages of reference-free metrics.
% Although the scores obtained using this approach do not approach the maximum score achievable with pseudocode as the ground truth (Table~\ref{tab
% }), we observe that the ranking of the models remains identical to those obtained when using the pseudocode as the ground truth. 
% We evaluated using two reference points: the GPT-generated pseudocode and the original protocol.
% G-Eval의 flexibility를 참고하여 ㄱㄱ
% 우리의 관점이 내포하는 바: original protocol이든 뭐든간에 ground truth로 사용할 수 있다. 
% (ADD above) Our results convey two potential interpretations.
% GPT관점에서 (gpt-percieved) protocol almost the same
% Second, 
% Although GPT is known to be prone to outputs of LLMs, pseudocodes may be a precise articulation in terms of identifying important information in scientific protocols. 
% (ADD should consider this as a preliminary study --)
%(ADD GPT-4 generated results are highest - >rescaling?) 
%(ADD cohere+ > GPT concern that GPT generated results may not be exactly the ground truth)
% 1. gpt pseudocode의 gpt-percieved quaility가 original protocol과 비슷하다. 
% (평가시 gpt가 attention하는 point가 pseudocode에도 잘 포함되어 있다.) 
% 2. 평가 방식에 있어서 비슷하다 > 평가할 때 원본 protocol을 주는 것을 고려
% new criteria를 define해서 얻는 장점은 oodi
\\\\
\noindent\textbf{Will LLM as an evaluator prefer responses from itself?}
It is reported that LLM as an evaluator prefer responses from itself over human responses in text summarization tasks~\citep{liu2023geval}.
Therefore, a potential concern is that the evaluator may prefer outputs from itself regardless of its quality.  While results in Table~\ref{tab:evaluator-comparison} and ~\ref{tab:results} address this concern, Table~\ref{tab:results-noaction} shows that Llama-3 as an evaluator does not prefer its outputs over that of GPT-4. Our results suggest that GPT's preference for its own responses in previous work~\citep{liu2023geval} may be a phenomenon unique to GPT.
% Already included above
% (ADD Llama-3 gave gpt-4 (baseline five points) which is compatible as a baseline)

% Our results potentially proposes the reason for GPT preferring responses from itself   maybe a phenomenon restricted to GPT.  

%여기에 bioprot 문제점 - introduction으로 옮김
\subsection{The \textsc{Bioprot 2.0} Dataset}
\begin{table}[h!]
  \centering
  % 공간없으면 작동
  \footnotesize
  \begin{tabular}{lc}
    \toprule
    Statistic & Value ($m\pm \sigma$)\\
    \midrule
    \# of protocols & 300 \\
    % Protocol length & $0 \pm 0$ \\
    Tokens / protocol & $812.3 \pm 469.9$ \\
    \# of steps & $14.81 \pm 10.74$ \\
    Tokens / step & $54.28 \pm 42.41$ \\
    Tokens / description  & $139.0 \pm 135.7$ \\
    Tokens / generated pseudocode  & $623.8 \pm 223.2$ \\
    \# of lines / generated pseudocode & $83.06 \pm 28.89$ \\
    \# of pseudofunctions / edited pseudocode & $10.28 \pm 6.582$ \\

% # of lines
% # of 슈도펑션

    % Tokens per generated description & $0 \pm 0$ \\
    \bottomrule
  \end{tabular}
  \caption{\textbf{Statistics of \textsc{Bioprot 2.0}.} `Edited Pseudocode' refers to the pseudocode that was reformatted, while preserving its content, to obtain the scores presented in Table~\ref{tab:results}. }
  \label{tab:pesudocode_stats}
\end{table}

% pseudofunction이 정확히 대응되지 않았다.
% GPT-4 as a annotator~\citep{finding_spoken_identifications_2024}
% step-by-step instruction generation (CoT)~\citep{li2024motion}
We introduce \textsc{Bioprot 2.0},  a dataset with scientific protocols and 
the corresponding pseudocodes with a larger number of datapoints.
Previous work highlights that a dataset with these two components can aid protocol formulation of LLMs~\citep{odonoghue-etal-2023-bioplanner}.
The pseudocode extracted from protocols are only composed of pseudofunctions (actions) predefined above the previous step, as each model was prompted to use only the provided functions but to define the arguments on their own. The summary of generated pseudocode are in Table~\ref{tab:pesudocode_stats}.
This dataset can be used to formulate scientific protocols to achieve a prompted goal using a toolformer like~\citep{schick2023toolformer} chain-of-thought LLM agent~\citep{wei2023chainofthought}. 

\section{Conclusion}
% Discussion 안쓰는 대신 results 가 analysis로 바뀜
%\section{Discussion}
% GPT 말고 다른 LLM도 evaluation에 사용될 수 있다. 

% Protocol의 수가 이전 연구에 비해서 많고, statistics도 비슷한 것으로 미루어 볼때, domain에 기반한 지식을 미리 반영하는 것은 성능향상을 이끌어낸다고 볼 수 있다. 

% 최근 연구와 compatiable 하게, traditional automatic metrics와는 차별화된 evaluation이 LLM으로 가능하다. 

% 아이디어
% Our framework은 
% 기존의 방법에서는 n-gram based metric으로 일괄적으로 평가를 진행하는 경우 target 모델이 오로지 슈도코드만이 아닌 다른 설명을 주석의 형태로 추가하여 실제로는 올바르게 함수를 지정했음에도 불구하고, 또는 의미론적으로는 같지만 정답지와 정확히 일치하지 않는다는 이유로 오답으로 처리되어 점수가 낮아지는 양상이 있었다. 
% (bioprot에서 볼 수 있었듯) 이러한 주석의 형태 등등이 추가되는 형상은 feedback loop을 적용해도 있었다. 
% G-Eval 써서 장점: 프로토콜 - 슈도코드 비교한 evaluation이 가능하다. (flexible)

% 기대효과
% This approach can be adapted for use in diverse fields, by modifying the protocol domain and implementing its knowledge accordingly.
% 무슨 capability 인지 설명적기? 

% 이 점이 다른분야 (화공)에 비해서 좋은 LLM의 등장을 막는 면도 있는 것 같다. 

% G-Eval 논문의 이점 언급해서 고찰
% without manual annotations to formulate baselines,
% transforming into format so that we can get traditional metrics/
% free of cost
% Our method enables relative comparison

% previous work에서 쓴 metric으로 구한 결과랑, g-eval이 이런면이 compatible 하다. 이런면이 낫다. 
% Fine-tuning해서 쓸 수 있는 bioprot 2.0의 의의 설명

We introduce \textit{ProtoMed-LLM}, a flexible and automatic framework designed to evaluate LLMs' capabilities on Scientific Protocol Formulation Tasks (SPFT). This framework prompts the target model and GPT-4 to extract pseudocode from biology protocols using only predefined lab actions, then evaluates the target model's output using \textsc{Llam-Eval}, with the GPT-4 generated pseudocode as a baseline and Llama-3 as the evaluator. Our prompt-based evaluation method, \textsc{Llam-Eval}, provides significant flexibility in terms of evaluation models, materials, criteria, and is free of cost. We assess various models, including GPT variants, Llama, Mixtral, Gemma, Cohere, and Gemini, and find GPT and Cohere to be particularly effective in formulating scientific protocols. Additionally, we present \textsc{Bioprot 2.0}, a dataset containing biology protocols and corresponding pseudocodes, which supports LLMs in the formulation and evaluation of SPFT. Our work is extensible to the assessment of LLMs on SPFT across various domains and other fields that require protocol generation for specific goals.

\clearpage
\section{Limitations}

% Notion에서 틈날때 작성

% --------------------------------------------------------------------
% 향후 글감

% Robustness
% (G-Eval limitation - 2 참고) LLM availabilty 에 전적으로 의존한다

% --------------------------------------------------------------------
We recognize several limitations. 
% 우리가 정의한 action이 biology lab에서 perform되는 모든 action을 포괄하지 못할수도 있다
The predefined actions may not encompass all actions performed in a biology labs.
% 불완전한 action deinement. Action을 정확하게 define 한다면 function에 더하여 function argument까지 define 해야한다. 
The definitions of predefined actions may be incomplete. To precisely define an action, it is necessary to define not only the function but also the function arguments.
% Evaluation을 위한 dataset으로서 Protocol 개수가 충분하지 않을 수 잇다. 
The number of protocols in \textsc{Bioprot 2.0} may be insufficient for evaluation purposes.
% biology라는 분야의 경계가 불명확해서 어디까지 (어떤 LLM까지) 쓰일 수 있는지가 불명확하다. 
The performance of \textit{ProtoMed-LLM} may decline outside of biology. Addressing this requires redefining domain-specific actions and exploring other LLMs for diverse fields. Future work should investigate these cross-disciplinary implications.
% GPT가 업그레이드 될 경우 metric의 수치가 이전모델을 사용한 것과 다를 수 있다. 
% GPT가 업그레이드 될 수 있고, 이논문에서 사용한 GPT-4가 사용 불가능해질수도 있다
LLMs are continuously evolving due to regular updates. The LLMs used for evaluation in this work might become unavailable in the future. Upgraded versions of LLMs may result in performance degradation and metrics may differ from those obtained using previous models.
% 우리는 GPT를 evlauator로 선택하였기 때문에 GPT의 bias, hallucination에 취약할 수 있다. 
% Future work은 다른 LLM을 evaluator로 해봐야한다.
% Llama말고 다른 모델로 evaluation했을 때는 어떻게 될지 모름
Due to selecting Llama-3 as the evaluator, our results may be susceptible to its biases and hallucinations. The outcomes when evaluated with models other than Llama-3 are unknown. Future work should investigate the outcomes using different LLMs as an evaluator. 
% (bioplanner limitation - 1 참고) GPT를 포함하여 evaluator로 API를 쓸때는 대부분 꽁짜가 아니며 비쌀 수 있다.
Using an API of LLMs as an evaluator, such GPT, is often not free of charge and can be costly.
% GPT-4 를 ground truth로 사용했지만 실제로 ground truth가 아닐 수 있다.
We used GPT-4 generated responses as a baseline; however, it may not accurately represent the ground truth. Future work should explore the implications of employing alternative resources (e.g., manually annotated pseudocodes, responses generated by other models) as the baseline. We observed basic actions classified as \verb|NoAction| in minor cases. 
%(G-Eval 언급내용) Will G-EVAL prefer LLM-based outputs? G-EVAL이 LLM-based output 선호
% (G-Eval 사용한 경우) bias towards LLM generated texts (G-Eval 논문 limitation 1) 
It has been reported that GPT prefers outputs from LLMs, which also produced our evaluation materials including all ground truth and target pseudocodes. This can potentially influence the scores.
%G-Eval에서 언급된 4가지의 특성이 real-world validation이 중요한 protocol의 evaluation역할을 충분히 못할 수도 있다.
The four criteria mentioned in G-Eval may not sufficiently fulfill the role of evaluating protocols where real-world validation is crucial. Also, applying these criteria originally designed for summarization tasks may be inappropriate for evaluating SPFT.
% Real-world Validation을 수행했지만 cell culture 보다 복잡하거나 (brain assembloids처럼) 섬세한 손작업이나 경험을 요하는 경우 프로토콜을 성공적으로 수행하였더라도 그 프로토콜을 수행하는 사람이나 물리적 장비의 조건에 따라 실험이 실패할 수도 있다. 
Even if the protocol pseudocode is successfully synthesized, real-world experiments may fail depending on the person performing the protocol or the condition of the physical equipment, especially in cases that are more complex than stem cell culture or require delicate manual work and experience.

\section*{Ethical Considerations}

% \noindent\textbf{Risk of Misuse} 
The use of manually verified protocols in LLMs is strictly prohibited for generating false protocols on platforms like STAR Protocols (Cell Press) and Nature Protocols. Numerous sites also prohibit the use of these protocols in conjunction with any form of AI tool. Our framework can be applied to the protocols of these sites. 
% (bioplanner limitation - 3 참고) Misuse 남용가능성
Although we have endeavored to exclude protocols that can create dangerous substances, there remains the potential for generating protocols that inadvertently produce hazardous products or byproducts.

% 잘못된 프로토콜을 생성할 수 있는 여지

\section*{Acknowledgements}
% 실험 도와주신분 감사인사
We would like to thank Karim Md.Adnan for assisting us with the action defining process. This research was supported by a KIST project (2E32351) and Bio-Cluster Industry Capacity Enhancement Project of Jeonbuk Technopark (JBTP) 

% Reference Pageover Switch
%\clearpage

\bibliography{anthology,custom}
\bibliographystyle{acl_natbib}

\clearpage
\section*{Appendix}

\appendix
\label{sec:appendix}

\section{\textsc{Bioprot 2.0}}

\subsection{Data Curation}
\label{sec:appendix-DatasetCuration}

We used \textit{protocols.io}~\citep{protocols.io} API for data collection. Protocols of $1 \leq score \leq 5$ and $3 \leq steps $ are collected. The collected data was in a \textit{.json} format, every data point with slight differences in keys. Some protocols were present in the git repository but could not be found when retrieved using the API, and vice versa\footnote{The protocol with ID 3737 exists in \textit{protocol.io} but doesn't exist in git repository.}. Also, even if the file ID in the git repository and the protocol ID retrieved using the API are the same, the dictionary key \textit{number\_of\_steps} may differ\footnote{The \textit{number\_of\_steps} for the protocol with ID 10489 is 3 in the git repository but 0 when retrieved using the API.}.
% key 어떻게 통일했는지 추가? 
% Scoring method 설명
 Keywords\footnote{The keywords are: Biology, Cell, DNA, Protein, Stem Cell, Molecular Biology, Molecular, Gene, Virus, E. coli, cDNA, Agarose, Agarose Gel, in vitro, PCR, NGS, Ethanol, Illumina, Cell Theory, Evolution, Genetics, Homeostasis, Cell Membrane, Mitochondria, Nucleus, Ribosomes, DNA Replication, Mutation, Chromosomes, Gene Expression, Natural Selection, Speciation, Adaptation, Phylogenetics, Ecosystems, Biodiversity, Conservation, Bacteria, Viruses, Fungi, Pathogens, Proteins, Enzymes, Metabolism, Photosynthesis, Gel Electrophoresis, Cloning, CRISPR-Cas9, Neurons, Brain, Synapses, Neurotransmitters, Antibodies, Vaccines, Immune Response, Autoimmunity, Embryogenesis, Stem Cells, Morphogenesis, Regeneration, Pollination, Growth Hormones, Tropisms, Coral Reefs, Oceanic Zones, Marine Conservation, Aquatic Ecosystems, Endangered Species, Habitat Destruction, Conservation Strategies, Rewilding, Genetic Engineering, Bioreactors, Bioinformatics, and Synthetic Biology.} extracted from the \textit{keywords.txt} file and the descriptions were converted to lowercase temporarily for comparison and scoring. As of May 2024, we collected a total of approximately 15k mirrored public protocols from \textit{protocols.io}'s GitHub before refinement.  Protocols were excluded if dictionary key \textit{steps} is empty. Protocols were manually verified by experts in biology. The protocols were removed if they were multiple duplicated files for an identical protocol\footnote{such as protocol ID: 9216}. For the same title, we score the latest version of the protocol.
 % Note that the statistics of the score was $00\pm00$ ($m\pm\sigma$).
% Keyword 어떻게 얻었는지는 일단 제외
% Keywords were obtained using ChatGPT, and those derived from brainstorming were classified under the Biology category.
% 숫자는 바뀔 수 있음
% Biology 주제와 관련된 프로토콜을 수집하기 위해 
% 다음과 같은 기준에 따라 프로토콜을 수집했다.
% \subsection{Data Refinement}
%\label{sec:appendix-Refinement}
% 아래 너무 Bioplanner랑 닮아있어서 용어변경 필요
% Collection할 때 word 점수 스코어링으로 할거면 filtering에서는 'no description 제외했다'는 배제 필요
% \\\\
% \noindent\textbf{Automatic Refinement}
%  Protocols were excluded if dictionary key \textit{steps} is empty.
% \begin{itemize}\itemsep-3pt
%     \item 
%     \item 2
%     \item Third item
% \end{itemize}
% (여기도) 아래 너무 Bioplanner랑 닮아있어서 용어변경 필요
% \noindent\textbf{Manual Refinement} Protocols were manually verified by experts in biology. The protocols were removed if they were multiple duplicated files for an identical protocol\footnote{such as protocol ID: 9216}.
% \begin{itemize}\itemsep-3pt
%     \item 
%     \item Second item
%     \item Third item
% \end{itemize}

% \subsection{Prompting Pseudocode Generation}
% \label{app:pseudocode-generation}
\subsection{Metrics and Evaluation}
\label{app:Metrics-Evaluation}
\textbf{Definitions of Evaluation Criteria}
\begin{itemize}\itemsep-3pt
    \item \verb|Consistency|: Consistency (1-5) - the factual alignment between the source and the target pseudocode. A factually consistent pseudocode contains only statements that are entailed by the source pseudocode. Annotators were also asked to penalize summaries that contained hallucinated facts.
    \item \verb|Fluency|: Fluency (1-5): the quality of the pseudocode in terms of grammar, spelling, punctuation, word choice, and structure.
    \item \verb|Relevance|: Relevance (1-5) - selection of important information from the source pseudocode. The target pseudocode should include only important information from the source document. Annotators were instructed to penalize summaries which contained redundancies and excess information.
    \item \verb|Precision|: Precision (1-5) - the exactness and accuracy of the expressions and terminology used in the pseudocode. The target pseudocode should avoid vague or ambiguous terms and should use specific and appropriate terminology that accurately reflects the intended operations and logic.
    \item \verb|Coverage|: Coverage (1-5) - the extent to which the target pseudocode addresses all aspects of the source pseudocode. The target pseudocode should comprehensively represent all the necessary steps, operations, and details present in the source pseudocode without omitting any critical information.

\end{itemize}
Note that above are criteria used for evaluation when GPT-generated pseudocode was a baseline. This was slightly modified when evaluating based on original protocol.
\\
\\
% 무슨 prompt 인지 설명
\noindent\textbf{Example \textsc{Llam-Eval} Prompt} Below is a prompt evaluating the generated pseudocode from a target LLM based on the criteria \verb|Coherence| using the GPT-generated pseudocode as the ground truth. The GPT-generated pseudocode for each protocol is placed inside \textit{\{\{Ground\_truth\_pseudocode\}\}}, and the target model-generated pseudocode is placed inside \textit{\{\{Target\_pseudocode\}\}}.
% .json이 여기여기 들어간다 간단 설명
\\\\
\textit{You will be given a source pseudocode as a ground truth. You will then be given a target pseudocode which is generated from an identical source of protocol.}

\textit{Your task is to rate the target pseudocode on one metric. Please make sure you read and understand these instructions carefully. Please keep this document open while reviewing, and refer to it as needed.}

\textit{Evaluation Criteria: Coherence (1-5) - the overall quality of all lines in the pseudocode. The target pseudocode should not be a rough overview but should provide a precise description of the ground truth pseudocode.}

\textit{Evaluation Steps:} 

\textit{1. Read the Ground Truth Pseudocode: Carefully read and understand the source pseudocode provided as the ground truth. Ensure you comprehend the logic, flow, and details of the algorithm or protocol described.}

\textit{2. Read the Target Pseudocode: Thoroughly read the target pseudocode that needs to be evaluated. Pay attention to the details, structure, and clarity of the pseudocode.}

\textit{3. Compare Against Ground Truth: Compare each line and section of the target pseudocode with the corresponding parts of the ground truth pseudocode.
Ensure that all critical steps, variables, and logic present in the ground truth are accurately reflected in the target pseudocode.}

\textit{4. Assess Coherence: Evaluate the overall quality of the target pseudocode based on how well it translates the ground truth. Consider the following aspects:
Clarity: Is the pseudocode easy to understand?
Completeness: Does it cover all the steps and details present in the ground truth?
Precision: Are the descriptions and instructions in the pseudocode precise and unambiguous?
Consistency: Are there any contradictions or logical inconsistencies?}

\textit{5. Assign a Coherence Rating (1-5):}

\textit{1 (Poor): The target pseudocode is incomplete, confusing, and lacks most details from the ground truth.
2 (Fair): The target pseudocode is partially complete but has significant gaps and is often unclear.
3 (Good): The target pseudocode covers most details from the ground truth but has some minor inconsistencies or lacks clarity in parts.
4 (Very Good): The target pseudocode is mostly complete and clear, with very few minor issues.
5 (Excellent): The target pseudocode is complete, clear, precise, and fully coherent with the ground truth.}
\\\\
\textit{Source Pseudocode: }

\textit{\{\{Ground\_truth\_pseudocode\}\}}
\\\\
\textit{Target Pseudocode: }

\textit{\{\{Target\_pseudocode\}\}}
\\\\
\textit{Evaluation Form (scores ONLY):}

\textit{- Coherence:}
% \\\\
% \noindent\textbf{SciBERTScore Definition}

\subsection{Evaluator LLM Selection}
\label{app:eval-llm-selection}
% self-self dataset 489, (GPT-4 = 488?) 

% (ADD larger dataset used)

% outliers excluded

Models without numerical responses include: Llama3-8b, Llama3-70b, Mixtral, and Gemma. 

\subsection{Implementation Details}
\label{app:implementation-details}
Except for \verb|n| and \verb|seed|, parameters were set to their default values. We used approximately \$1000 for GPT API calls, \$20 for Gemini, and other models were free of cost. 
% \noindent\textbf{Parameters} Except for \verb|n| and \verb|seed|, were set to their default values. 
% \\\\
% \noindent\textbf{Budget}
% We used approximately \$1000 for GPT API calls, \$20 for Gemini, and other models were free of cost. 
%\\\\
%\noindent\textbf{}
\\\\
\noindent\textbf{Counting Tokens} We counted the tokens of the concatenated string of the title, original description, and steps, separated by "$\backslash$n$\backslash$n". The reason for this approach is to match the token count with that of the previous work.
\\\\
\noindent\textbf{Inconsistencies \textsc{G-Eval} Outputs} 
To address this issue, we attempted the following methods: (1) Modified $\verb|max_token|=5$ to  $\verb|max_token|=1$ : The scores became integers, but the model still generated sentences in addition to scores. (2) Use different versions of the model: Other model variations, such as \verb|gpt-3.5-turbo-1106|, did not enhance the results. 
% \begin{itemize}\itemsep-3pt
%     \item Modified $\verb|max_token|=5$ to  $\verb|max_token|=1$ : The scores became integers, but the model still generated sentences in addition to scores.
%     \item Use different versions of the model: Other model variations, such as \verb|gpt-3.5-turbo-1106|, did not enhance the results. 
%     % \verb|gpt-4o|: 우리가 위에 쓴거라서.. 보류
%     % \item Use edited data set: produces more consistent results.
%     % --------- (여기서부터 작성) ---------------
%     % \item Use edited prompt: Edited prompts according to xx provides better results than the original prompt.
%     % \item Reduce the text tokens we provide.
%         % \begin{itemize}\itemsep-3pt
%         %     \item It seems to be more effective than previous attempts.
%         % \end{itemize}
% \end{itemize}
% \\\\
% \noindent\textbf{Packages}

% in self-self we manually excluded scores >5 or expectionally large values (outliers)

% \\\\
% default values (geval, gpt) description separately
% \\\\
% Non-digit responses were excluded.
% \\\\
% The specific versions used for each models can be found at Table~\ref{tab:version_stats}.
% Bioplanner가 suffle 1번으로 나와서 순서 변경? 

%Citation List
%Galactica \citep{taylor2022galactica} \\
%PubMedQA \citep{jin-etal-2019-pubmedqa} \\ 
%GPT-3.5 (?논문은 안오는데)

\subsection{Evaluating LLMs on SPFT}
\label{app:eval-llms-spft}

\textbf{Versions of LLMs}
\begin{table}[htbp!]
  \centering
  % 공간없으면 작동
  % \small
  \begin{tabular}{lc}
    \toprule
    \textbf{Model Name} & \textbf{Call Strings} \\
    \midrule
    GPT-4o & \verb|gpt-4o|\\
    GPT-4 & \verb|gpt-4| \\
    GPT-3.5 & \verb|gpt-3.5-turbo-1106| \\
    Llama3-8b & \verb|llama3-8b-8192| \\
    Llama3-70b & \verb|llama3-70b-8192| \\
    Mixtral & \verb|mixtral-8x7b-32768| \\
    Gemma-7b & \verb|gemma-7b-it| \\
    Cohere+ & \verb|command-r-plus|\\
    Cohere & \verb|command-r| \\
    Gemini-1.0 & \verb|gemini-1.0-pro-001| \\
    Gemini-1.5& \verb|gemini-1.5-pro-001| \\
    Gemini-2.0& \verb|gemini-1.0-pro-002| \\
    \bottomrule
  \end{tabular}
  \caption{\textbf{Versions of LLMs.} Exact API call strings for corresponding models.}
  \label{tab:version_stats}
\end{table}

\end{document}